\documentclass[10pt,journal,letterpaper,compsoc]{IEEEtran}

\usepackage{graphicx}
\usepackage{amsmath,amssymb,longtable}
\usepackage{color}
\usepackage[T1]{fontenc} 
\usepackage{float}       
\usepackage{helvet}      
\usepackage{mathptmx}    
\usepackage{setspace}    
\usepackage{url}         
\usepackage{algorithmic}
\usepackage{algorithm}

\title{An iterative feature selection method for GRNs inference by exploring topological properties}

\author{
Fabr\'{\i}cio M. Lopes\thanks{F.M. Lopes is with the Federal University of Technology - Paran\'a, Brazil (e-mail: fabricio@utfpr.edu.br).},
David C. Martins-Jr\thanks{D.C. Martins-Jr is with the Federal University of ABC, Brazil (e-mail:david.martins@ufabc.edu.br),},
Junior Barrera\thanks{J. Barrera is with the Faculty of Philosophy, Sciences and Letters of Ribeir\~{a}o Preto, University of S\~{a}o Paulo, Brazil (e-mail: jb@ime.usp.br).} and
Roberto M. Cesar-Jr\thanks{R.M. Cesar-Jr is with the Institute of Mathematics and Statistics of the University of S\~ao Paulo and Brazilian Bioethanol Science and Technology Laboratory (CTBE), 
Brazil (e-mail: cesar@vision.ime.usp.br)}
}

\begin{document}
\maketitle

\begin{abstract}
An important problem in bioinformatics is the inference of gene regulatory
networks (GRN) from temporal expression profiles.
In general, the main limitations faced by GRN inference methods
is the small number of samples with huge dimensionalities and the noisy nature of
the expression measurements.
In face of these limitations, alternatives are needed to get better accuracy
on the GRNs inference problem.
This work addresses this problem by presenting an alternative feature
selection method that applies prior knowledge on its search strategy, called SFFS-BA.
The proposed search strategy is based on the Sequential Floating
Forward Selection (SFFS) algorithm, with the inclusion of
a scale-free (Barab\'{a}si-Albert) topology information in order to guide
the search process to improve inference.
The proposed algorithm explores the scale-free property by pruning
the search space and using a power law as a weight for reducing it.
In this way, the search space traversed by the SFFS-BA method combines
a breadth-first search when the number of combinations is small
($\langle k \rangle \leq 2$)
with a depth-first search when the number of combinations becomes
explosive ($\langle k \rangle \geq 3$), being guided by the scale-free prior information.
Experimental results show that the SFFS-BA provides a better inference similarities
than SFS and SFFS, keeping the robustness of the SFS and SFFS methods,
thus presenting very good results.
\end{abstract}
\section{Introduction} 
\label{sec:intro}
One of the most challenging research problems for System Biology nowadays is the inference
(or reverse-engineering) of gene regulatory networks (GRNs) from expression profiles.
This research issue became important after the debioinformaticsvelopment of high-throughput technologies for
extraction of gene expressions (mRNA abundances or transcripts), such
as DNA microarrays \cite{shalon1996} or SAGE \cite{velculescu1995},
and more recently RNA-Seq \cite{wang2009}.
This problem regards revealing regulatory relationships between biological molecules in order to
recover a complex network of interrelationships, which can describe not just diverse biological
functions but also the dynamics of molecular activities.
It is very important to understand how many biological processes happen and in most cases,
how to prevent it from happening (diseases).

In the context of expression profiles, a big challenge that researchers
need to face is the large number of variables or genes
(thousands) for just a few experiments available (dozens).
In order to infer relationships among those variables, it is needed a
great effort in developing novel computational and statistical techniques
that are able to alleviate the intrinsic error estimation committed in
the presence of small number of samples with huge dimensionalities.

In general, it is not possible to recover the GRNs very accurately.
The main reasons for this are the lack of information about the biological organism,
the high complexity of the networks and the intrinsic noise of the expression measurements.
In this context, there are several recent initiatives to overcome such limitations by
incorporating other information on the inference/prediction method.
One kind of initiative is the use of the functional gene information, e.g., from the Gene Ontology, Proteome, KEGG,
among others, into the clustering process, resulting in more biologically meaningful clusters
\cite{macintyre2010,cui2010,dehaan2010}.
Another alternative is by using biological information for the discovery of
transcriptional regulation relationships, i.e., to infer GRNs \cite{werhli2008, ernst2008, seok2010}.
A variety of biological data integration techniques for GRNs inference are described in
\cite{troyanskaya2005,karlebach2008,baumbach2009,hecker2009}.

Although the integration of biological information with mathematical models is critically important
in discovering novel biological knowledge, it is restricted by the
prior biological information of each gene or biological entity.
One way to use prior information and make the methods less restrictive is 
the use of information about local or global prior knowledge of an organism 
instead of an information about a single gene, e.g.,
to use the network structure/topology as prior information.
In this way, the integrated use of multiple data types together with local and global topological properties
could be decisive for the effective prediction of GRNs and their functions in face of the known limitations
\cite{troyanskaya2005,vidal2005,aittokallio2006,ray2009,kuchaiev2010}.

The analysis of local and global biological network properties
and its application on the inference process is very recent and promising
\cite{klamt2007,lacroix2008,karlebach2008,lenas2009,przytycka2010}.
For example, the application of network structure by the inference methods
makes use of similarities of connected network modules \cite{ulitsky2007},
structural and graph-theoretic interpretation for the network components
\cite{narasimhan2009,kuchaiev2010}, taking into account the network sparseness
\cite{andrecut2008,andrecut2008b,christley2009}, gene network motifs \cite{ott2005}
and the search for cliques in network graphs \cite{yu2006}.
More important, biological networks and particularly GRNs are known
not only to be sparse, but also organized, so as nodes belonging to
different connectivity classes \cite{charbonnier2010}.
These examples show the importance of such problem and the need for
new methodologies to overcome it.

The information about network topology can help the investigation of
biological process by adopting the complex network theory and its properties
\cite{strogatz2001,albert2002,newman2003,costa2008}.
It is known that many kinds of relationships can be successfully described by complex networks.
In particular, the complex networks theory describes various types of network topologies,
each one with well defined properties, and it has been broadly applied to
characterize biological processes and gene relationships involved. Some biological networks,
for example, were shown to present the scale-free property, in which many nodes have a low
degree and a few of them have a high degree (hubs), in which the degree distribution
is approximated by a power-law distribution
\cite{jeong2000, guelzim2002, farkas2003, barabasi2004, albert2005, costa2008, barabasi2009}.
In general, topological patterns and its structural analysis is one of the most promising
topics in the analysis of complex networks \cite{alon2007,goemann2009}, and particularly
the application of structural properties of the network can be a very valuable prior
information to be used by the GRNs inference methods.

The use of network topology models for the simulation and analysis of GRNs have been recently 
described in \cite{lopes2011a} and the same models are further explored in the present paper. 
In this work, a new method is proposed for the inference of GRNs from expression profiles
by incorporating a scale-free topology and applying a conditional dependency criterion
function. Thus, it is suggested a method that takes the scale-free topology into
account as prior information in the inference process.
This leads to a better inference of networks presenting scale-free property.
The main purpose of this paper is to show the interest of taking into account
information about the scale-free topology of the network in order to
improve the inference process and make it more suitable for a class of problems, 
i.e., scale-free networks.

The inference process is conducted by observing the conditional dependence of a
target gene given its potential predictors through temporal expression profiles,
and by applying the mean conditional entropy as criterion function
\cite{liang1998,martins2006,barrera2007,lopes2008b}.
This process has been recognized as an appropriate
statistical tool to model direct interactions between genes \cite{charbonnier2010}.

The main contribution of this work is the proposal of a new feature
selection method for GRNs inference from temporal expression profiles.
Our inference method is based on a previous feature selection
algorithm \cite{pudil1994, lopes2010},
with the inclusion of scale-free topology as a prior information,
in which the search space traversed is relatively small and provides
encouraging results.

Next sections (Sections \ref{sec:complexnet} and \ref{sec:inference})
introduce a brief background on the complex network theory and the network
inference problem.
In Section \ref{sec:featureselection},
the feature selection problem is discussed in more detail, including a
short description of the SFS and SFFS techniques.
Section \ref{sec:imp} discusses the intrinsically multivariate prediction
issue and how it can affect the greedy feature selection algorithms in
such way that the achieved solution may be relatively far away from the optimal.
Section \ref{sec:sffsba} describes our proposed feature
selection method (SFFS-BA). Section \ref{sec:results} shows some
experimental results. Finally, Section \ref{sec:conclusion} concludes the
work, discussing future perspectives.

\section{Complex Networks Theory and Biological Networks}
\label{sec:complexnet}
The genes of a network can be characterized by its degree, i.e., the number of
connections with other genes of these network that it has.
By considering directed networks, there are two kinds distinct relationships.
The in-degree is the number of directed connections received by a gene.
The out-degree is the number of directed connections edges sent by a gene. 
The individual gene degrees can be used to estimate the degree distribution $P(k)$ 
and as a result, characterize the whole network, i.e., a global network property.

The uniformly-random Erd\"{o}s-R\'{e}nyi (ER) \cite{er1959} complex network model
is based on randomly connected vertices.
This model assume the hypothesis that complex systems are connected at random,
leading to a Poisson degree distribution with peak at the average degree $\langle k \rangle$, 
indicating that the most of the genes have a degree close to $\langle k \rangle$.
In other works, the ER model has a statistically homogeneous degree distribution \cite{albert2005}.

On the other hand, the scale-free network structure proposed 
by Barab\'{a}si and Albert \cite{ba1999} (BA), is based on a heterogeneous distribution
on its vertex degree, in which few genes have a large number of connections and the most
of genes have few connections. The absence of a typical degree led to this complex
network model to be described as ``scale-free'' \cite{albert2005}.
More specifically, the scale-free structural property is characterized by a power-law in 
its connections (degree) distribution.
In other words, the probability $P(k)$ of a gene to interact with $k$ other genes 
decays as a power law

\begin{equation}
    P(k) \sim k^{- \gamma} \text{,}
    \label{eq:powerlaw}
\end{equation}

\noindent in which $\gamma$ is a numerical constant.

In general, numerous networks, such as the Internet, human collaboration
networks and metabolic networks, follow a scale-free structure \cite{albert2005}.
In particular, most known biological networks present a scale-free structure \cite{costa2008},
implying that their distribution on its gene relationships (degree $k$) is irregular,
a large number of connections (edges) is concentrated on a small number of genes,
while large number of genes have few connections.
In this case, scale-free networks have a high probability of exhibiting hubs \cite{ba1999}.

In particular, by considering the transcriptional regulations of the \textit{Saccharomyces cerevisiae}
\cite{guelzim2002}, it was found a close relation between proteins and genes which presents a degree
distribution with an exponential decay very similar to a power law.
In the work developed by Farkas \textit{et al.} \cite{farkas2003}, it was found an overlap
between the connectivity distribution of scale-free and \textit{Saccharomyces cerevisiae}
transcriptome networks. In such work it was suggested a potential regulatory relationship among its genes, 
in which a small number of transcription factors are responsible for a complex set of expression 
patterns under diverse conditions.

Regarding the constant decay $\gamma$, it was reported that scale-free networks describe 
the \textit{Escherichia coli} metabolic networks and its metabolic reactions follows a power-law, 
with $\gamma = 2.2$ \cite{jeong2000}.
It is also known that the probability of a given yeast protein to interact with $k$ other 
yeast proteins follows a power-law, with $\gamma = 2.4$ \cite{jeong2001,boccaletti2006}.
In general, the degree exponent $\gamma$ is usually in the range
$2 < \gamma <  3$ \cite{albert2002,albert2005}.
In summary, the scale-free complex network model has been effectively used to simulate and 
describe the behavior of biological networks
\cite{costa2008, jeong2000, guelzim2002, farkas2003, barabasi2004, albert2005, barabasi2009}.

\section{GRN Inference}
\label{sec:inference}
The combination of expression analysis, perturbations, treatments and
gene mutations may indicate information about molecular or specific
functions of the genes. Gene regulatory networks (GRN) inference from
expression data, a process also known as reverse engineering, is a
difficult computational task due to the huge data volume (number of
genes or expression profiles) and to the small number of available
samples, including the large complexity of biological networks,
thus representing an important challenge in bioinformatics and
computational biology researches \cite{hovatta2005}.

The GRNs inference from temporal expression data tries to identify the
variation of the expression levels along time, becoming possible
to indicate information such as metabolic pathways, cell cycle and
mapping of modifications caused by stimuli. It can be used as a model
for functional representation of gene interactions.

It is important to highlight that the network inference has the
objective of discovering interaction networks between genes that are
potentially interesting from the biological point of view. The
relationships among genes are suggested according to some estimator
and can be examined or validated by wet-lab experiments. As such
experiments have a high cost in terms of financial, human and time
resources, the main idea is to offer to the specialists a reduced
number of hypotheses that satisfactorily identify a certain phenomenon
of interest.

There are several approaches for modeling and identification of GRNs. 
Examples include Boolean Networks \cite{kauffman1969} and its stochastic 
version (Probabilistic Boolean Networks) \cite{shmulevich2002pbn},
Differential Equations \cite{chen99} and Bayesian Networks \cite{friedman2000},
to name but a few.
This work focuses on the Probabilistic Boolean
Networks (PBN) model, since it captures global properties of GRNs
while dealing well with settings presenting limited number of samples.
                                                                    
Regarding the feature selection approaches to infer GRNs, there are mainly
three types of criterion functions. The correlation based criterion functions
are those that measure 1-to-1 relationships, often employed to identify
co-regulation between genes, functional modules and clusters \cite{stuart2003}.
It does not take into account multivariate relationships, i.e., the expression
of a given target being regulated by a set of two or more genes with
multivariate interaction.
Bayesian error estimation based criterion functions evaluate the estimated
errors present in the joint probability distribution of a target gene given
its candidate predictor genes \cite{hashimoto2004,doughertybrun2007,ghaffari2010}, 
which is capable to detect multivariate (N-to-1) relationships. 
Finally, information theory based criterion functions are used to detect 1-to-1
and N-to-1 relationships \cite{margolin2006,faith2007,rao2007,barrera2007,zhao2008}, 
relying on the uniformity of the conditional probability distributions of the
target given the candidate predictors as a whole (larger uniformity leads to
higher entropy, which in turn leads to smaller mutual information).

The literature related to modeling and identification of GRNs is huge and on
rapidly increasing, which indicates its importance. The reader is referred to
\cite{karlebach2008,hecker2009,dhaeseleer2000,jong2002,styczynski2005,schlitt2007}
for reviews on this subject.

\section{Feature Selection}
\label{sec:featureselection}
Pattern recognition methods allow the classification of objects, i.e., 
a class or label is assigned to each object based on its features. In many
applications, and specifically in GRN inference, the feature space dimension
of such objects tends to be very high while
the number of samples is very limited. In this
context, the study and development of techniques for dimensionality
reduction become mandatory.

Feature selection is a possible approach to perform dimensionality
reduction. A feature selection method looks for subsets of features that lead
to a good representation, classification or prediction of the objects
classes. It is composed by two main parts: a search algorithm and a criterion
function which assigns a quality value to the feature subsets.

The only way to guarantee optimality of the solution is to investigate the
entire space of possible subsets (exhaustive search), although depending on
some criterion function constraints adopted (e.g. monotonical or U-shaped), it
is possible to reach the optimal solution by searching for a constrained
subset space by applying ``branch-and-bound'' methods \cite{somol2004,ris2010}.
The exhaustive search is computationally unfeasible in general, and especially
for inference of GRNs which involves data with thousands of features (genes).
There are many heuristics proposed for feature selection. Two classical
feature selection heuristics are briefly introduced and discussed below.

\subsection{Sequential Forward Selection (SFS)}
\label{ssec:sfs}
The Sequential Forward Selection (SFS) is a genuinely greedy feature selection
algorithm in the sense that it includes the best feature according to the
criterion function in each step. It starts with an empty set
(bottom-up approach) and adds the best
individual feature to the partial solution. In the next step, it looks for the
best feature that, jointly with the feature already included in the partial
solution, forms the best pair. This process continues until a stop condition
is satisfied, which normally is based on a fixed dimension value (cardinality
of the subset to be returned) or based on the variation of the criterion
function from the previous to the next step. A variant of this algorithm is
the Sequential Backward Selection (SBS) which starts with the full feature set
and eliminates the least relevant feature according to the criterion function
(top-down approach),
repeating this process successively until a stop condition is satisfied
\cite{pudil1994}.

The greedy algorithms such as SFS and SBS present a drawback known as nesting
effect. This effect occurs because the discarded features by using the
top-down approach are never inserted again to the partial solution,
and the inserted features in the bottom-up
approach are never discarded from the partial solution.
Section~\ref{sec:imp} presents the reason why this phenomenon occurs.

\subsection{Sequential Floating Forward Selection (SFFS)}
\label{ssec:sffs}
\begin{figure}[h!]
\begin{center}
  \includegraphics[width=\linewidth]{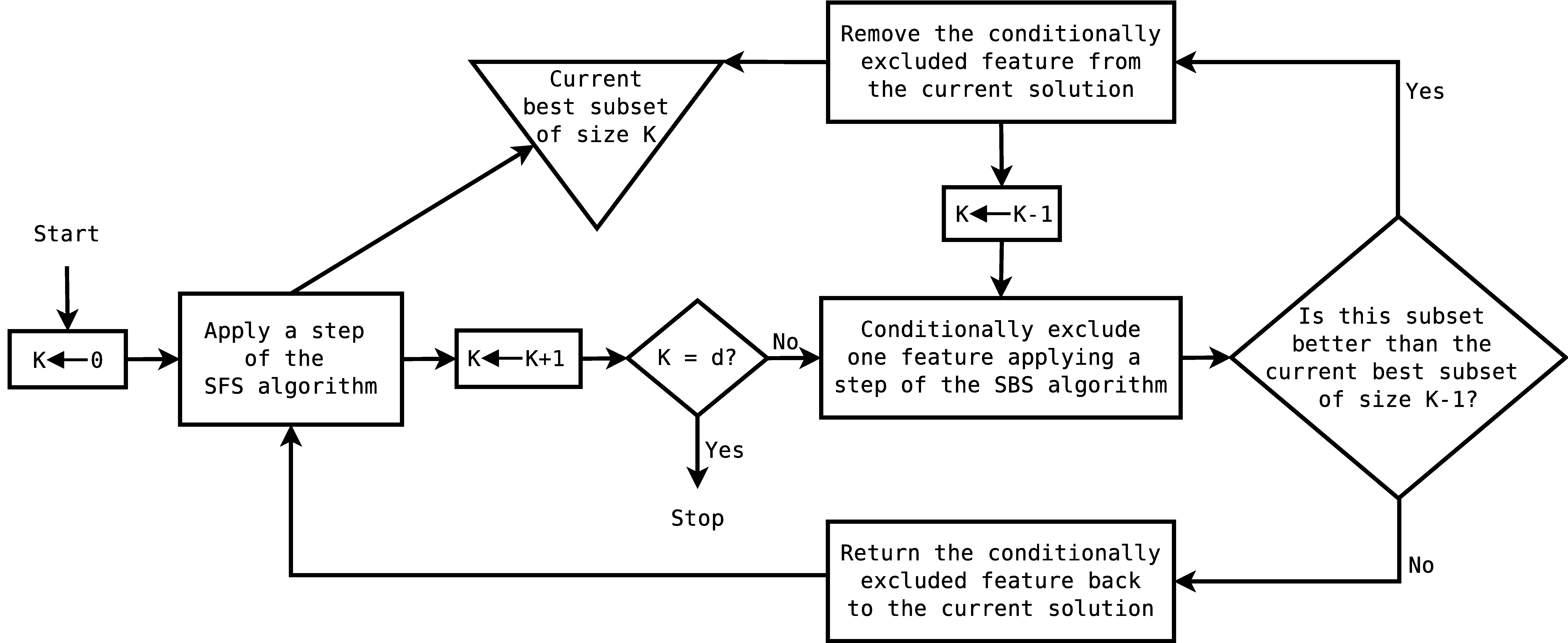}
  \caption{Schematic flowchart of the SFFS algorithm \cite{lopes2008b}
  (adapted from \cite{somol1999}).
   $K$ refers to the size of the current solution subset while $d$ refers
   to the size of the final solution subset.}
\label{fig:sffs}
\end{center}
\end{figure}

The Sequential Floating Forward Selection (SFFS) algorithm tries to alleviate the nesting
effect by allowing the inclusion and exclusion of features in the partial solution in a
floating way, i.e., without requiring the definition of the number of
insertions or exclusions \cite{pudil1994}. It starts with an empty set
(cardinality $k = 0$). The SFS algorithm is applied until $k = 2$. For $k > 2$, 
the SBS algorithm is applied in order to exclude bad features. The SFFS
applies alternately the SFS and SBS until a stop condition is reached. The
best solution of each cardinality $k$ is stored in a list. The best solution
among them is selected as the returned solution of the algorithm and, in case
of ties, the solution with the smallest cardinality is returned. A schematic
flowchart of the SFFS algorithm is presented in Figure \ref{fig:sffs}.

SFFS is renowned for presenting an excellent cost-benefit in terms of the
computational complexity and the quality of the returned solution.
There are some variants of this algorithm (adaptive and generalized floating
search methods) that try to improve the SFFS solutions at the expense of an
increase on the computational cost.
However, they can not avoid the nesting effect completely \cite{somol1999}.

\subsection{Intrinsically Multivariate Prediction}
\label{sec:imp}
This section briefly discusses one of the main reasons why the feature
selection heuristics do not guarantee the optimum solution. A target feature
is intrinsically multivariate predicted (IMP) by its predictor feature set if
all predictors combined greatly predicts the target behavior, while every
properly contained subset of the mentioned predictor set has an almost null
prediction power regarding the target. Formally, a set of features
$\mathbf{X}$ is
intrinsically multivariate predictive for the target feature $Y$ with
respect to $\lambda$ and $\delta$, for $0 \leq \lambda ,\delta \leq 1$
and $\lambda < \delta$, if

\begin{equation}
\label{eq:imp}
\max_{\mathbf{Z}\varsubsetneqq \mathbf{X}}\mathcal{F}_{Y}(\mathbf{Z})\:\leq\:
\lambda \wedge \mathcal{F}_{Y}(\mathbf{X})\geq \delta
\end{equation}
\noindent where $\mathcal{F}$ is a criterion function that varies from 0
to 1 (0 meaning absence of prediction and 1 meaning full
prediction) \cite{martins2008}. The parameters $\lambda$ and $\delta$ usually
assume small (less than 0.2) and large (greater than 0.8) values,
respectively, to reflect the IMP
property. Additionally, an intrinsically multivariate
predictiveness degree (IMP score) through the maximum value of
$\delta - \lambda$ can be defined as:

\begin{equation}
I_{Y}(\mathbf{X})=\mathcal{F}_{Y}(\mathbf{X})-\max_{\mathbf{Z}\varsubsetneqq
\mathbf{X}}\mathcal{F}_{Y}(\mathbf{Z})
\label{eq:impness}
\end{equation}

Considering Boolean features, the deterministic exclusive-or binary logic
(XOR) is an example of an IMP set when two predictors may assume any value
from $\{(0,0), (0,1), (1,0), (1,1)\}$ with uniform probability distribution.
In this case, it is
impossible to predict the target based on the observation of just one of its
predictors, since the target can assume the value 0 or 1 regardless of the
values of each individual predictor. Of course when the predictors are
combined, the prediction is perfect for every instance from
$\{(0,0), (0,1), (1,0), (1,1)\}$ (the prediction is given by the XOR logic).

The nesting effect occurs in most feature selection algorithms and can be
explained by the IMP phenomenon. Considering an IMP set, its individual
features (or subsets of features) are not good to predict the target, so they
hardly will be included in the partial solution of a given sub-optimal feature
selection. However, an optimum solution
can be formed by such features together (large IMP score), which implies that
the considered
feature selection probably does not reach this solution. Besides, two
good individual predictors may not lead to a very good pair, since they have a
relatively high correlation with the target feature, meaning that they have
high correlation with each other. Section~\ref{sec:sffsba} shows that the
method proposed here can eventually return IMP sets as solutions when its IMP
score is moderately large, i.e., each individual feature has little, but not
null, contribution in predicting the target feature, and also when the
cardinality of such IMP sets is not very high (in this case, any good
predictor set with large cardinality would not be returned due to the
high estimation error performed when evaluating large dimensionality sets).

\section{Probabilistic Genetic Networks}
\label{sec:pgn}
Probabilistic Genetic Networks (PGN) is a model proposed to represent
GRNs. PGNs are based on PBNs, in which the selection of the transition
function is not deterministic and the states of the genes and networks are
represented by discrete values. PGNs describe a finite dynamical system,
discrete in time, composed by a finite number of states, in which each
transcript is represented by a variable. The composition of all variables form
a vector considered the system state. Each vector component has an associated
transition function which calculates its next value from the previous state of
other genes (predictors). These functions are components of a transition
functions vector, which defines the transition from a network state to the
next state and represents the gene regulation mechanism \cite{barrera2007}.

In this model, the gene expression networks are represented as a stochastic
process ruled by a Markov chain. In other words, assuming this principle means
that a conditional probability of a future event, given the previous events,
depends only on the immediately previous event. A Markov chain is
characterized by a transition matrix $\pi_{Y|X}$ of conditional probabilities
among states, and its elements are denoted by $p_{y|x}$, and a vector of
initial states $s_0$. A PGN is a Markov chain $(\pi_{Y|X},s_0)$ in which some
axioms are assumed \cite{barrera2007}:

\begin{itemize}
\item[i] the transition matrix $\pi_{Y|X}$ is homogeneous, i.e., $p_{y|x}$ is
not a function of $t$. The state transition probability is constant;

\item[ii] $p_{y|x} > 0$, i.e., all state pairs can be reached
(ergodic Markov chain);

\item[iii] the transition matrix $\pi_{Y|X}$ is conditionally independent,
      i.e., for every state pair $x, y$, $p_{y|x} = \prod_{i=1}^n p(y_i|x)$;

\item[iv] $\pi_{Y|X}$ is quasi-deterministic, i.e., for every state $x$,
exists a state $y$ such that $p_{y|x} \approx 1$.
\end{itemize}

Theses axioms are motivated by biological phenomena or simplifications due to
the lack of samples for model estimation. The first axiom is a constraint to
simplify the estimation problem. The second axiom states that all states are
reachable, i.e., the presence of perturbation or noise can eventually lead the
system to any state. The third axiom determines that a gene expression at a
given time instant is independent of the expression of other genes at the same
instant $t$. The last axiom says that the system has a structural dynamics
which is prone to small noises \cite{barrera2007}.

\section{SFFS with Structural Properties (SFFS-BA)}
\label{sec:sffsba}
%
%
The proposed method is based on the Probabilistic Genetic Networks (PGN),
which is described in \cite{barrera2007} and in Section~\ref{sec:pgn}.
The proposed method considers the four axioms established by the PGN model
and proposes two new constraints:

\begin{itemize}
  \item For every target gene $i$, by adding a new predictor $x_i$ in the result set with cardinality $> 1$,
    there should be an information gain in the prediction of the target gene, whenever the chosen predictor
    is part of the true result set. If the information gain by adding a new predictor in the result set is poor,
    the predictor could not be part of the true result set or there is no data enough to make the prediction.
    In both cases, the inclusion of the predictor in the result set should be avoided.
  \item The network topology follows a power-law in its connections distribution, i.e.,
    a scale-free network structure such as described in Section~\ref{sec:complexnet}.
\end{itemize}

By assuming the PGN model and these new constraints, the main contribution of
our method is to include structural information as a prior knowledge to
perform a search on a reduced space, thus achieving better results.

\begin{figure}[ht!]
    \centering
    \includegraphics[width=0.95\linewidth]{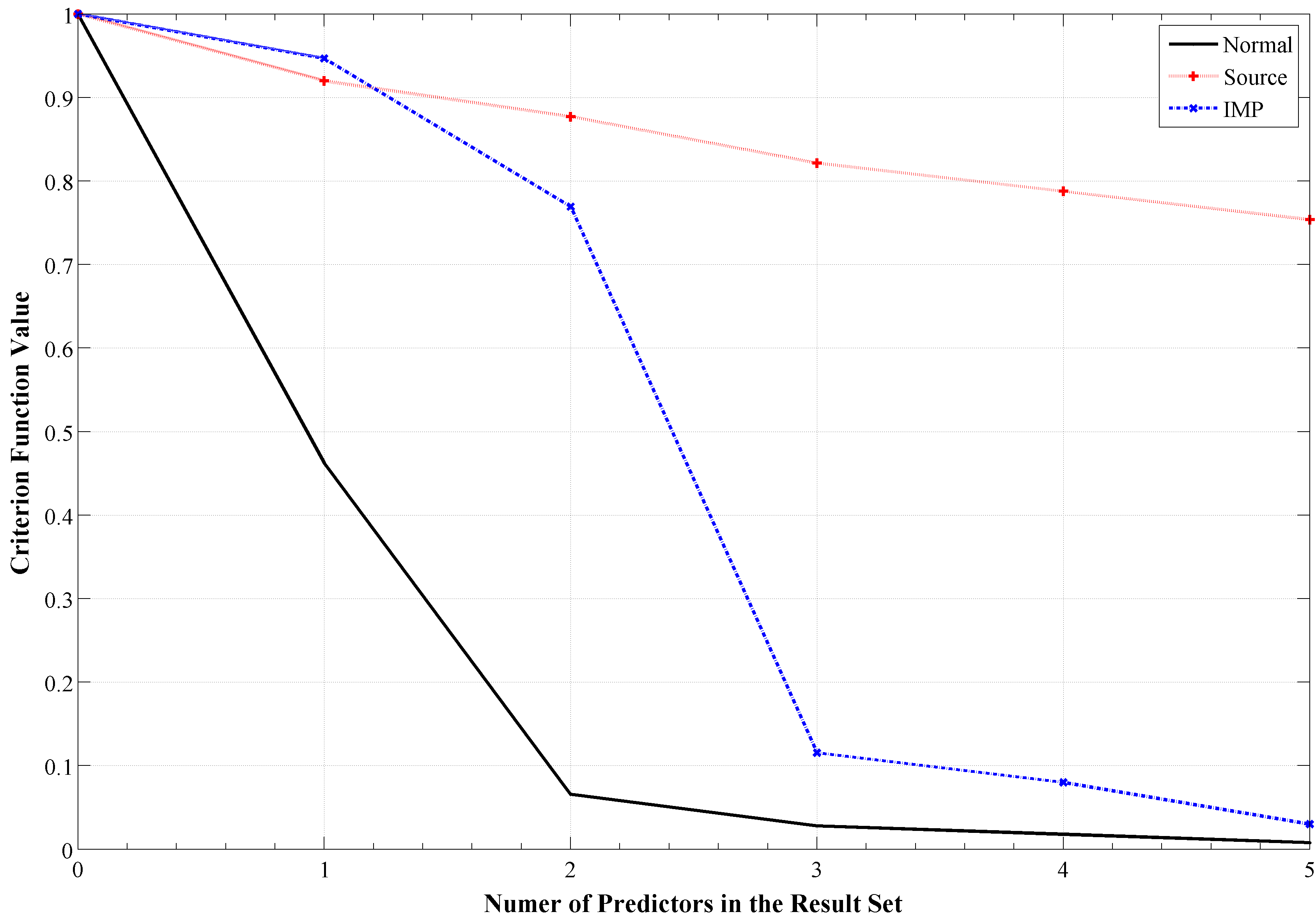}
    \caption{Criterion function behavior by the inclusion of new predictors in
    the result set (optimum value is zero). The black curve (Normal) is expected 
    for targets with well defined predictors in which each predictor has a good
    contribution in predicting the target behavior. The blue curve (IMP) is expected
    for targets with intrinsically multivariate predictors in which each predictor 
    is not good enough to predict the target behavior, as opposite to the whole predictor set. 
    The red curve (Source) is expected for targets that have no predictors.}
    \label{fig:cfvariation}
\end{figure}

The idea is based on the assumption that a gene with no predictors tends to have a 
random behavior, while a gene with predictors tends to have a more ordered behavior.
In this way, it is possible to expect that a source gene in a GRN
(i.e., with no predictors) presents a behavior with small variations in the 
criterion function on trying to identify a possible predictor to it.
In other words, small variations are expected on the criterion function values
by adding new predictors on the result set of a source, as shown by
Figure \ref{fig:cfvariation} (Source).

On the other hand, when a gene has predictors, it is expected a distinct
behavior, mainly on trying a possible predictor that is part of the result set
(true positive).
In this way, it is expected some significant variation on criterion function
values when performing
a search for possible predictors, as showed by Figure \ref{fig:cfvariation}
(Normal and IMP).

The curves shown in Figure \ref{fig:cfvariation} were obtained using the 
AGN model described in \cite{lopes2011a}, for which it was applied the mean
conditional entropy as a criterion function \cite{lopes2008b} in order to exemplify
the adopted assumption for the three discussed cases, i.e., normal, source and IMP. 

Given the network topology constraint and the IMP property, the proposed algorithm
performs a search for best and worst individual predictors,
based on the SFFS-MR algorithm \cite{lopes2010}, by considering all genes of
the network. SFFS-MR is an algorithm that applies the classical SFFS for
several initial features, considering good and bad individual features.
In the second step, the search is performed again for all genes, but the
algorithm chooses the target genes
that present a prediction gain (i.e., the criterion function value moves
closer to the optimum) by adding a new predictor to its result set.
This increased result set is preserved in the next iteration.
The target genes with small information gain, i.e., poor predictions when 
increasing its predictor set, are not considered for the next iteration, 
as well as the predictor set that reaches the optimal value
of the criterion function or gets too close to it.

In this context, the scale-free topology and the above presented assumption
are considered by the proposed algorithm in order to integrate the prior
knowledge (scale-free topology), applied to reverse-engineering of GRNs
from temporal expression profiles.
By considering that the scale-free network model is characterized by a
power law $P(k) \approx k^{-\gamma}$ in its connections distribution
(Section~\ref{sec:complexnet}), the same power law is considered to
prune the search space on each iteration.

\begin{algorithm}[!ht]              
  \caption{Network-Inference ($targets$, $exps$, $\gamma$, $\Delta$)}   
  \label{alg:inference}               
  \begin{algorithmic}[1]            
      \STATE \textbf{var} \textit{list}  $exelist$
      \STATE \textbf{var} \textit{integer} $k \leftarrow 1$, $n \leftarrow targets$.size()
      \FOR{$i = 1$ to $n$}
	  \STATE $exelist$.append($targets[i]$, $\emptyset$, $1$, $0$)
      \ENDFOR
      \WHILE{$n > 1$}
	  \FOR{$i = 1$ to $n$}
	    \STATE $[target,psets,cfv,gain]$ $\leftarrow$ $exelist$.removefirst()
	    \STATE $[newpsets,newcfv,gain]$ $\leftarrow$ SFFS-BA($target$,$cfv$,$psets$,$k$,$exps$,$\Delta$)
	    \STATE exelist.append($target$,$newpsets$,$newcfv$,$gain$)
	  \ENDFOR
	  \STATE SortPredictorSetsbyGain(exelist)	
	  \STATE $n \leftarrow n \times k^{-\gamma}$
          \STATE $k \leftarrow k + 1$
      \ENDWHILE
      \RETURN $exelist$
  \end{algorithmic}
\end{algorithm}

\begin{algorithm}[!ht]              
  \caption{SFFS-BA ($target$,$cfv$,$psets$,$k$,$exps$,$\Delta$)}   
  \label{alg:sffsba}               
  \begin{algorithmic}[1]            
      \IF{$psets = \emptyset$}
	\FOR{$predictoridx = 1$ to $exps$.size()}
	    \STATE $psets$.append($predictoridx$)
	\ENDFOR
      \ENDIF
      \WHILE{$psets$ is not empty \AND $psets$.first.cardinality $\leq k$}
	\STATE $newpset$  $\leftarrow$ $psets$.removefirst()
	\STATE $newcfv$ $\leftarrow$ SFS($target$,$newpset$,$k$,$exps$)
	\IF{$newcfv$ $<$ $bestcfv$ \AND $(bestcfv-newcfv)$ $> \Delta$}
              \STATE $newcfv$  $\leftarrow$ SBS($target$,$newpset$,$exps$)
              \STATE $bestcfv$ $\leftarrow$ $newcfv$
	      \STATE $bestset$ $\leftarrow$ $newpset$
        \ENDIF
	\IF{$newpset$.cardinality = 1}
	      \STATE $psets$.append($newpset$)
	\ENDIF
      \ENDWHILE
      \IF{$k > 1$}
	\STATE $psets \leftarrow bestset$
      \ENDIF
      \RETURN [$psets$, $bestcfv$,$(cfv-bestcfv)$]
  \end{algorithmic}
\end{algorithm}

Algorithm \ref{alg:inference} starts by considering the targets and all available
samples (temporal expression profile, called \textit{exps}) in order to select the
individual predictors, i.e., $k=1$. 
In the following, Algorithm \ref{alg:sffsba} (SFFS-BA) is applied in order to
discover the best features of each target gene, which are ranked according to
the adopted criterion function. 
One important difference of the proposed feature selection method is that the algorithm
will explore the search space in steps, i.e., the predictors are chosen iteratively
according to the cardinality parameter $k$.
Another difference is that for $k=1$ the SFFS-BA algorithm will return all predictor
sets and the best criterion function value in order to explore all the individual
predictors in the next iteration and to better recover the predictors of the IMP
targets.
From $k > 1$, the algorithm begins to return only the best set, assuming that
some of the true predictors would be in the selected predictors set.

At the end of each iteration of the Algorithm \ref{alg:inference},
the target genes are sorted by the prediction gain,
the number of considered targets for the next iteration is updated following
a power law, given by $n = n \times k^{-\gamma}$ and the cardinality of the
result set is updated ($k \leftarrow k+1$).

In this way, when $k=2$, the next iterations will consider just the target genes
with higher prediction gain when increasing its predictor cardinality $k$.
It is important to notice that target genes that reach the optimal value of
criterion function or get too close are not considered for the increasing on
its predictor cardinality.
The search is performed while the number of target genes $n > 1$ (stop condition).

In summary, the SFFS-BA differs from SFFS (Section \ref{ssec:sffs}) because of its
iterativity, the exploration of all combinations of predictors set with cardinality 
$k \leq 2$ and the inclusion of a search space pruning method based on the assumption
that the expression data (input) were generated from a scale-free network.

Algorithms \ref{alg:inference} and \ref{alg:sffsba} present the specification of
the proposed feature selection algorithm: SFFS-BA.

The parameter $exps$ represents the temporal expression profile, in which
the genes are generally arranged in the rows and experiments in the columns.
The parameter $\gamma$ is a constant value that determines the exponential decay,
i.e., the number of targets that will be considered in the next iteration
(predictor set expansion).
A criterion function value variation ($\Delta$) is also included.
The $\Delta$ value prevents that minor
variations of the criterion function ($\leq\Delta$) cause
the increase of the predictors subset.
The present paper adopted $\gamma = 2.5$ and $\Delta = 0.05$.
The adopted $\gamma$ is related with the mean value found in the literature,
which
is usually in the range $2 < \gamma < 3$ \cite{albert2002,albert2005}.

The application of the algorithm to predict a single gene or a set of genes
of interest instead of the entire network is straightforward by selecting the
targets parameter.

\section{Experimental Results}
\label{sec:results}
This section presents the experimental results obtained by considering
a synthetic networks approach, which is described in \cite{lopes2011a}.
The artificial gene networks (AGNs) were generated by considering the
uniformly-random Erd\"{o}s-R\'{e}nyi (ER) topology,
the scale-free Barab\'{a}si-Albert (BA) and 
the small-world Watts-Strogatz (WS).

For all experiments, the network models (ER, BA and WS) were applied with
$n=100$
vertices (genes). The average degree $\langle k \rangle$ per gene
varied from 1 to 5, and the number of observed time instants
(signal size) varied from 5, 10, 15, 20 to 100 in steps of 20.
For each gene $g_i$ of the network, its value was given by
a randomly selected function from 3 possible Boolean functions
$\{f_{1}^{(i)},f_{2}^{(i)},f_{3}^{(i)}\}$, which represents
different behaviors or functions assumed by each gene $g_i$.
In order to assign a robust structural dynamics with small noise to the
networks, the probabilities of each function be selected are given by
$c_{1}^{(i)}=0.98,\ c_{2}^{(i)}=0.01,\ c_{3}^{(i)}=0.01,\ i=1,\ldots,n$. With
these probabilities, the PGN axioms ii (all possible states are reachable) and
iv (quasi-deterministic setting) are satisfied (Section~\ref{sec:pgn}).

The network identification method described in \cite{lopes2008b}
implements feature selection methods for network inference.
By applying the SFS and SFFS as search strategies and the mean
conditional entropy as criterion function.
This method was applied in order to identify the networks
from simulated temporal expressions.
The same method, criterion function and other parameters (default)
were kept fixed during the comparative analysis with SFFS-BA.

In order to measure the similarity between the synthetic ($A$) and the
inferred ($B$) networks,
we adopt the $PPV$ (Positive Predictive Value, also known as accuracy or precision) and
$Sensitivity$ (or recall) \cite{dougherty2007,lopes2011a}, which are widely used to
compare the results of the GRNs inference methods.
Since the $PPV$ and $Sensitivity$ are not independent of each other, we take into account
the geometrical mean between them as a similarity measure,
given by: $Similarity(A,B) = \sqrt{PPV(A,B) \times Sensitivity(A,B)}$.

\begin{figure}[ht!]
    \centering
    \begin{tabular}{cc}
        \includegraphics[width=0.8\linewidth]{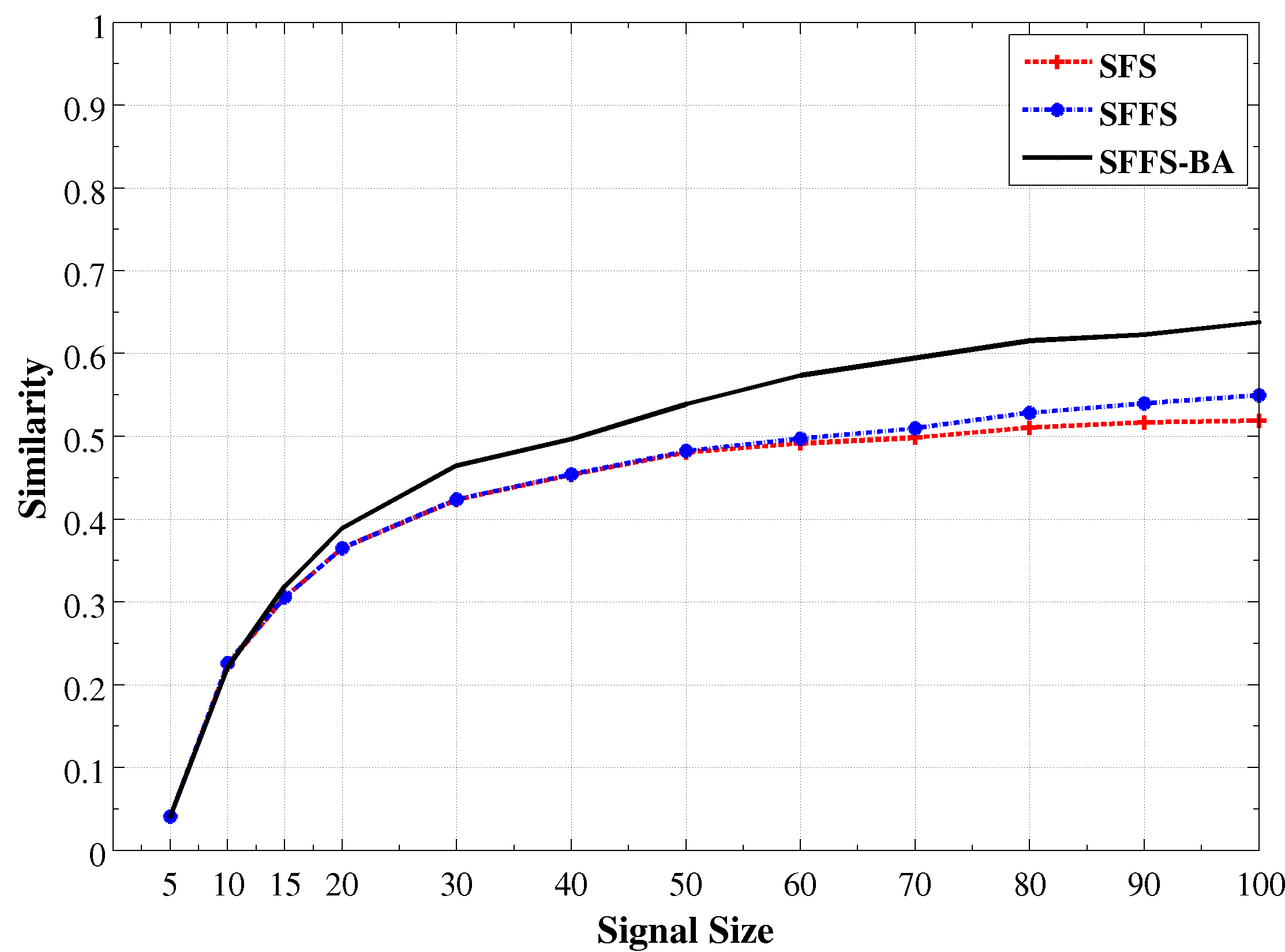} & \raisebox{0.310\linewidth}{(a) ER}\\ 
        \includegraphics[width=0.8\linewidth]{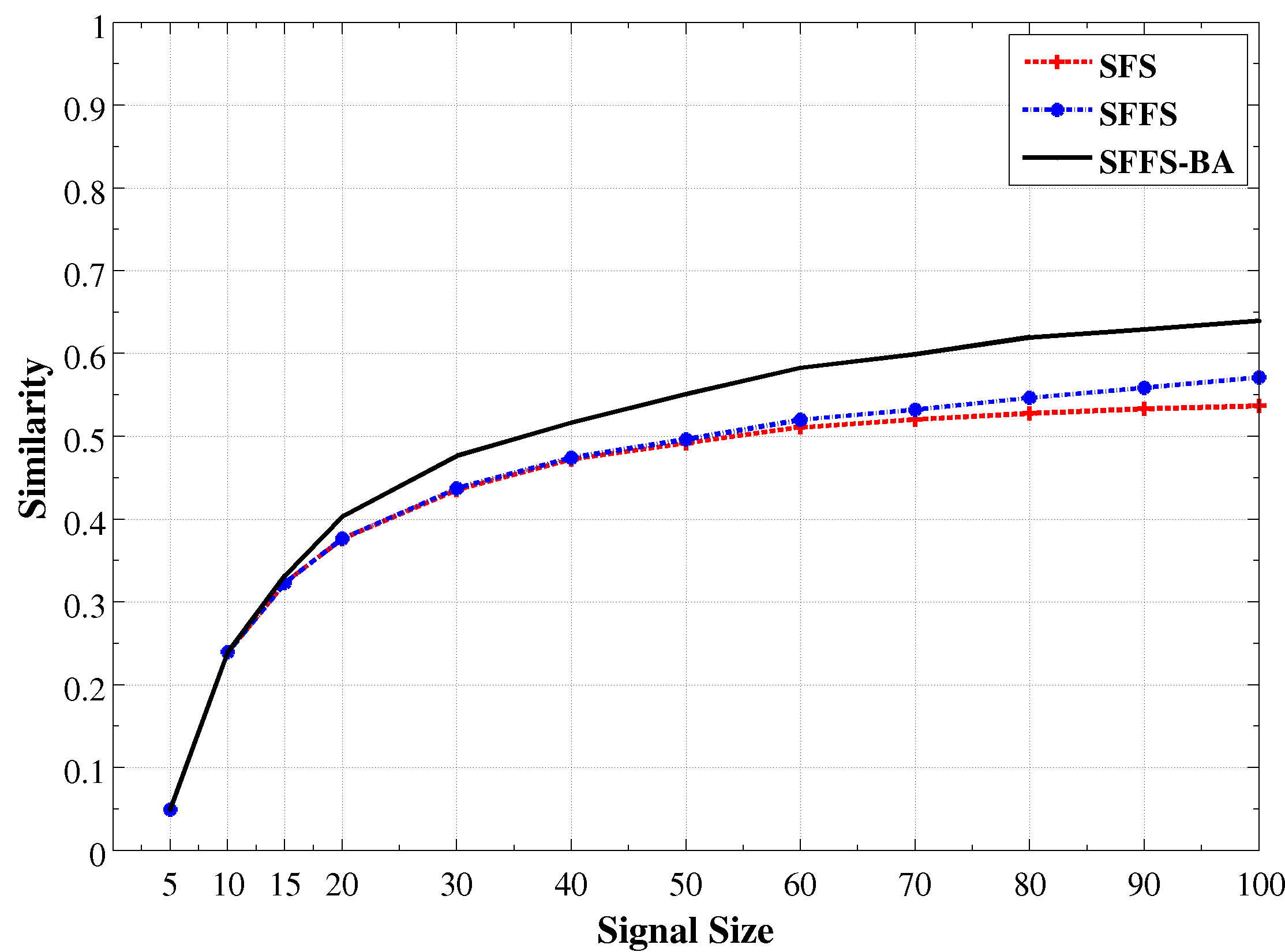} & \raisebox{0.310\linewidth}{(b) WS}\\
        \includegraphics[width=0.8\linewidth]{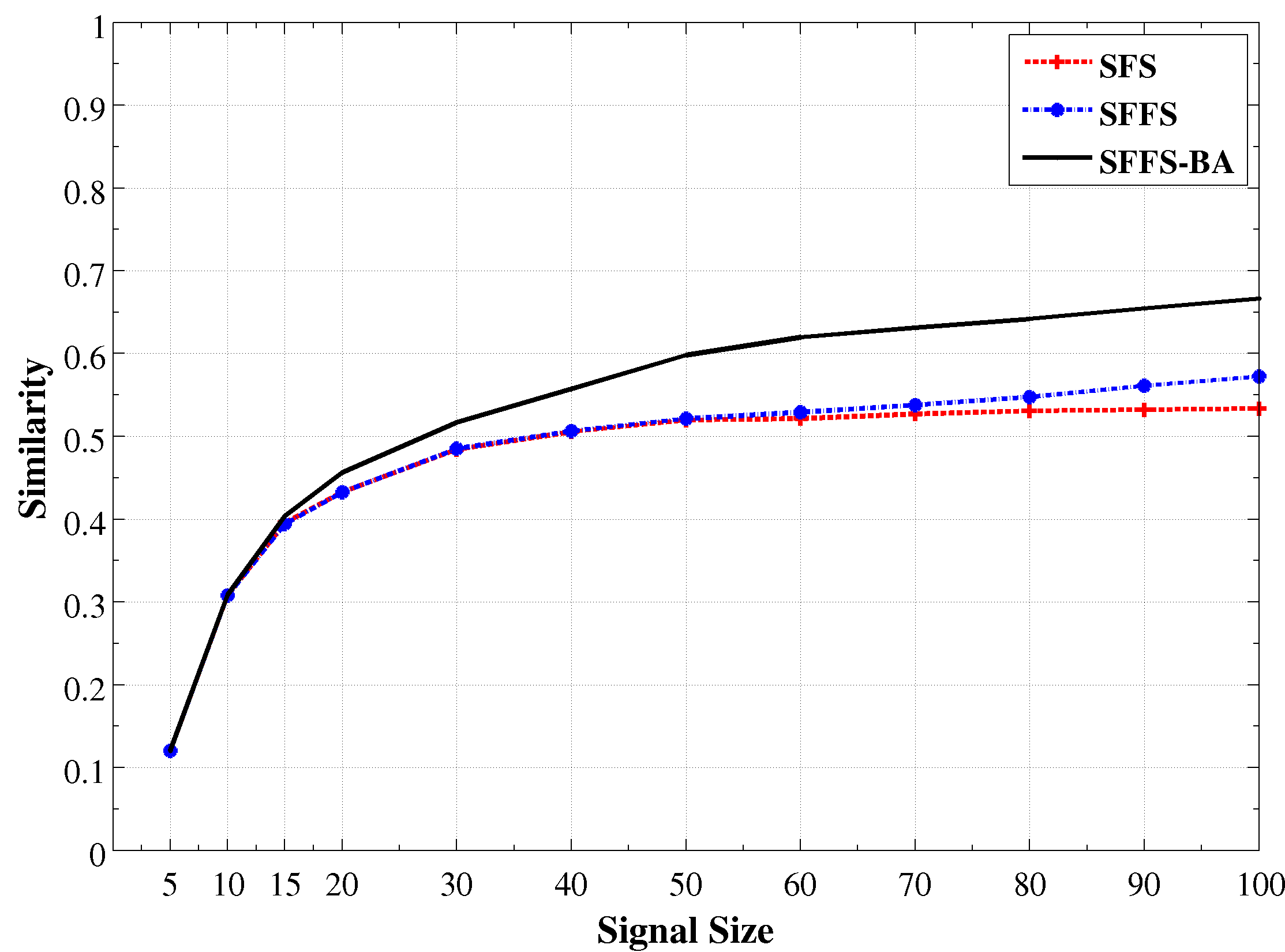} & \raisebox{0.310\linewidth}{(c) BA}\\
    \end{tabular}
    \caption{Similarity measure obtained by SFS, SFFS and SFFS-BA
	    applied to infer network edges from temporal expression profiles
	    with different number of time points (signal size).
            The similarity represents the mean over 50 executions for each network topology.}
    \label{fig:signalsize}
\end{figure}

The experimental results were obtained from 50 simulations
by using different signal sizes (i.e., number of time points) and 
$\langle k \rangle$ values.
The first experiment was performed in order to compare
the three methods (SFS, SFFS and SFFS-BA) with respect
to the temporal expressions size,
which is a critical issue in bioinformatics problems.
Figure \ref{fig:signalsize} presents these results,
in which the similarity measure was calculated by taking
into account the average results for all variations of $\langle k \rangle$.

\begin{figure}[ht!]
    \centering
    \begin{tabular}{cc}
        \includegraphics[width=0.8\linewidth]{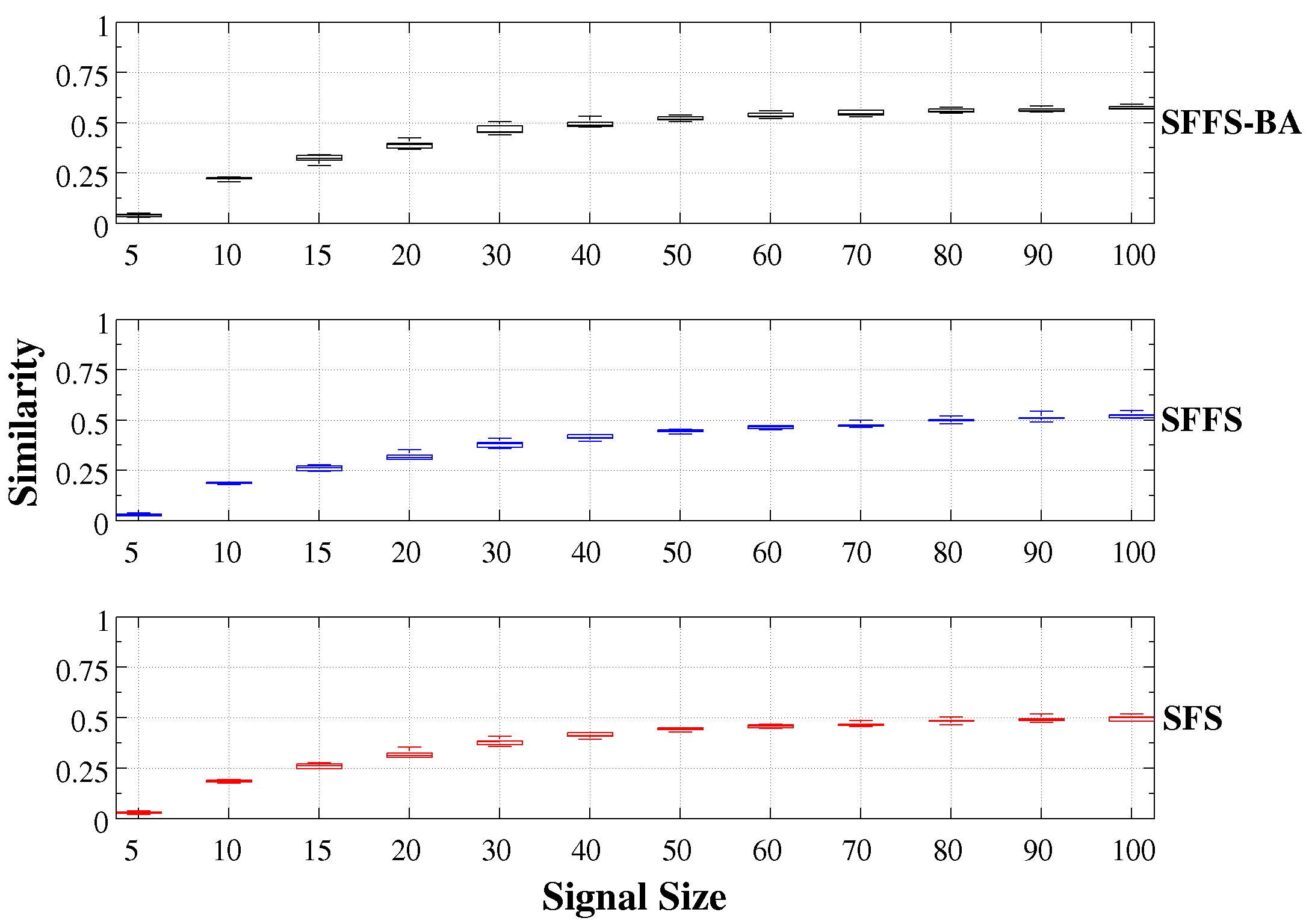} & \raisebox{0.295\linewidth}{(a) ER}\\ 
        \includegraphics[width=0.8\linewidth]{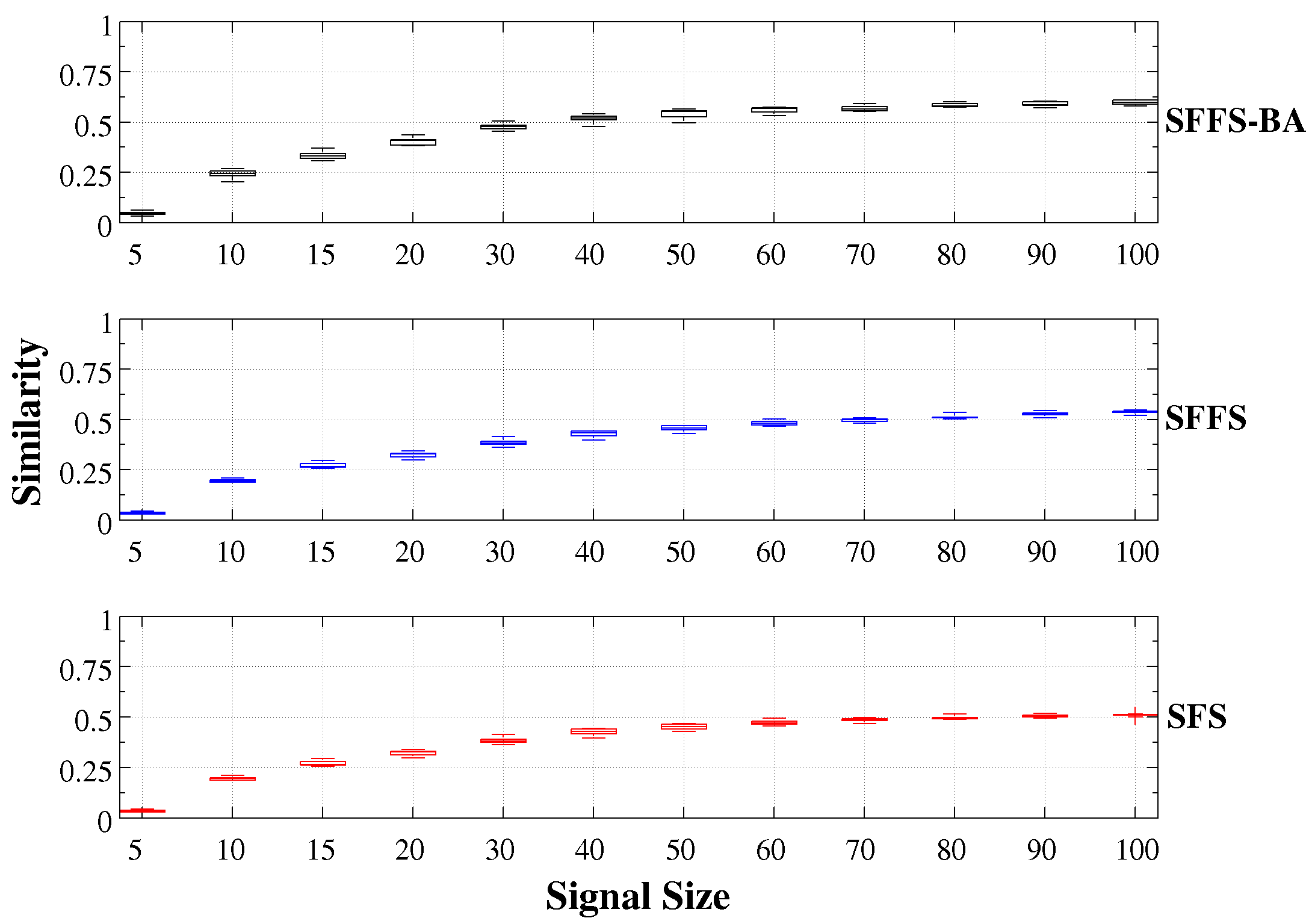} & \raisebox{0.297\linewidth}{(b) WS}\\
        \includegraphics[width=0.8\linewidth]{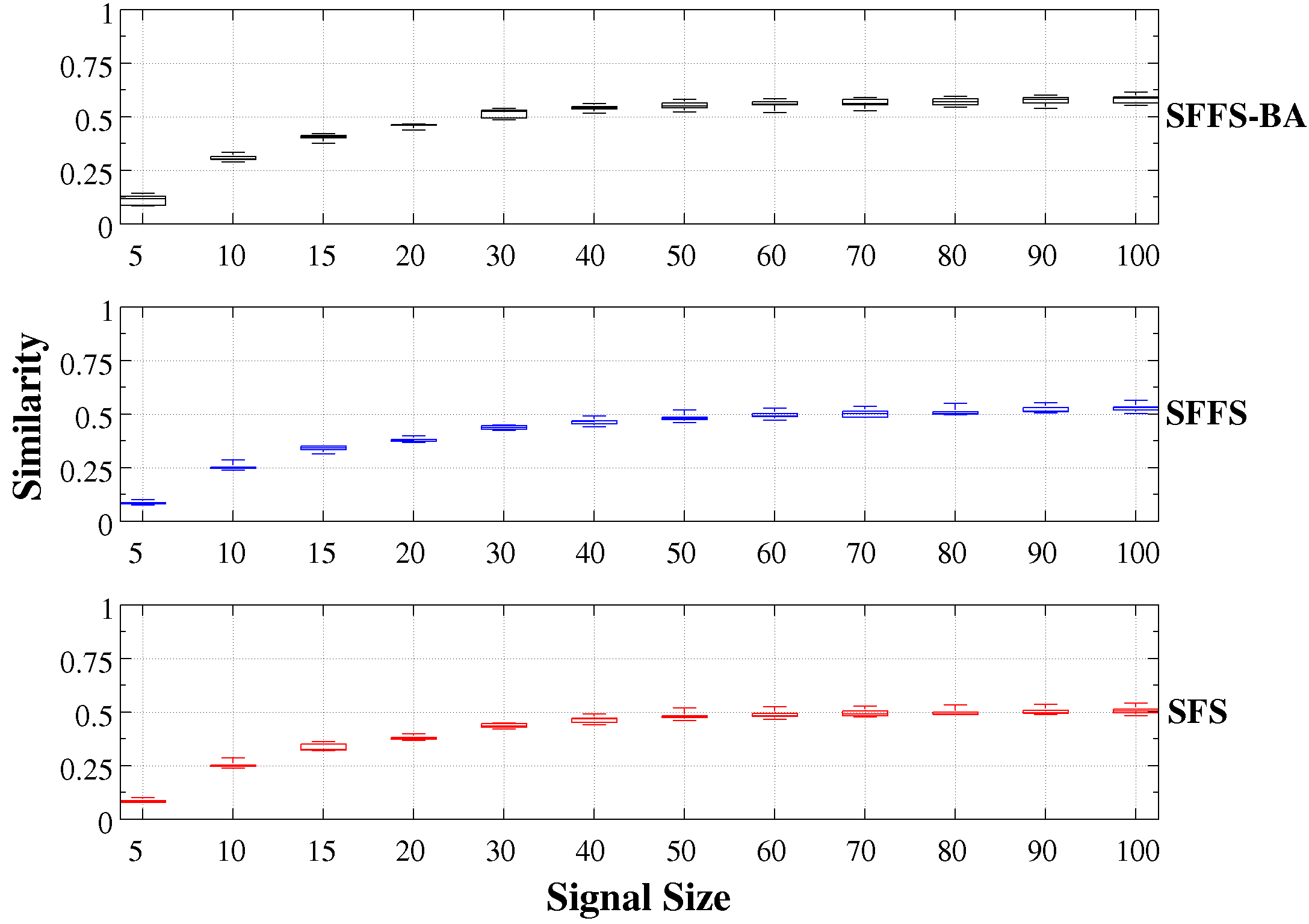} & \raisebox{0.3\linewidth}{(c) BA}\\
    \end{tabular}
    \caption{Distribution (boxplots) of the similarity values obtained by SFS, SFFS and SFFS-BA
	    applied to infer network edges from temporal expression profiles
	    with different number of time points (signal size) from 50 executions of each network topology.}
    \label{fig:signalsizeboxplots}
\end{figure}

It is possible to notice that all methods have an increase on
its performance by increasing the number of observations in the
three network topologies (ER, WS and BA).
However, the improvement of the SFFS-BA occurs earlier
by using just 20 time points, consistently outperforming the other two methods
from this point forward.
Meanwhile, the SFFS slightly outperforms the SFS only with the signal size
greater than 80 time points, i.e., the difference
of the similarity rates is smaller than the difference achieved by SFFS-BA
by considering all network topologies.

In addition, the SFFS-BA similarity curve (Figure \ref{fig:signalsize})
shows a more significant improvement with the expansion on signal size
by considering the BA network topology, as expected.
However, considering ER and WS network topologies, the improvement of the
SFFS-BA
not only outperforms the other two methods but also it is consistent, even in the
presence of some perturbations in the temporal signal, which is implied by the
stochasticity in the application of transition functions.
Figure \ref{fig:signalsizeboxplots} (a) and (b) shows the boxplots of the similarity 
values for each number of time points.
It is possible to notice a very small variation in the boxplots, 
indicating stable results for all time points.
These results are an important indicative of the stability of
the proposed methodology.

\begin{figure}[ht!]
    \center
    \begin{tabular}{cc}
        \includegraphics[width=0.8\linewidth]{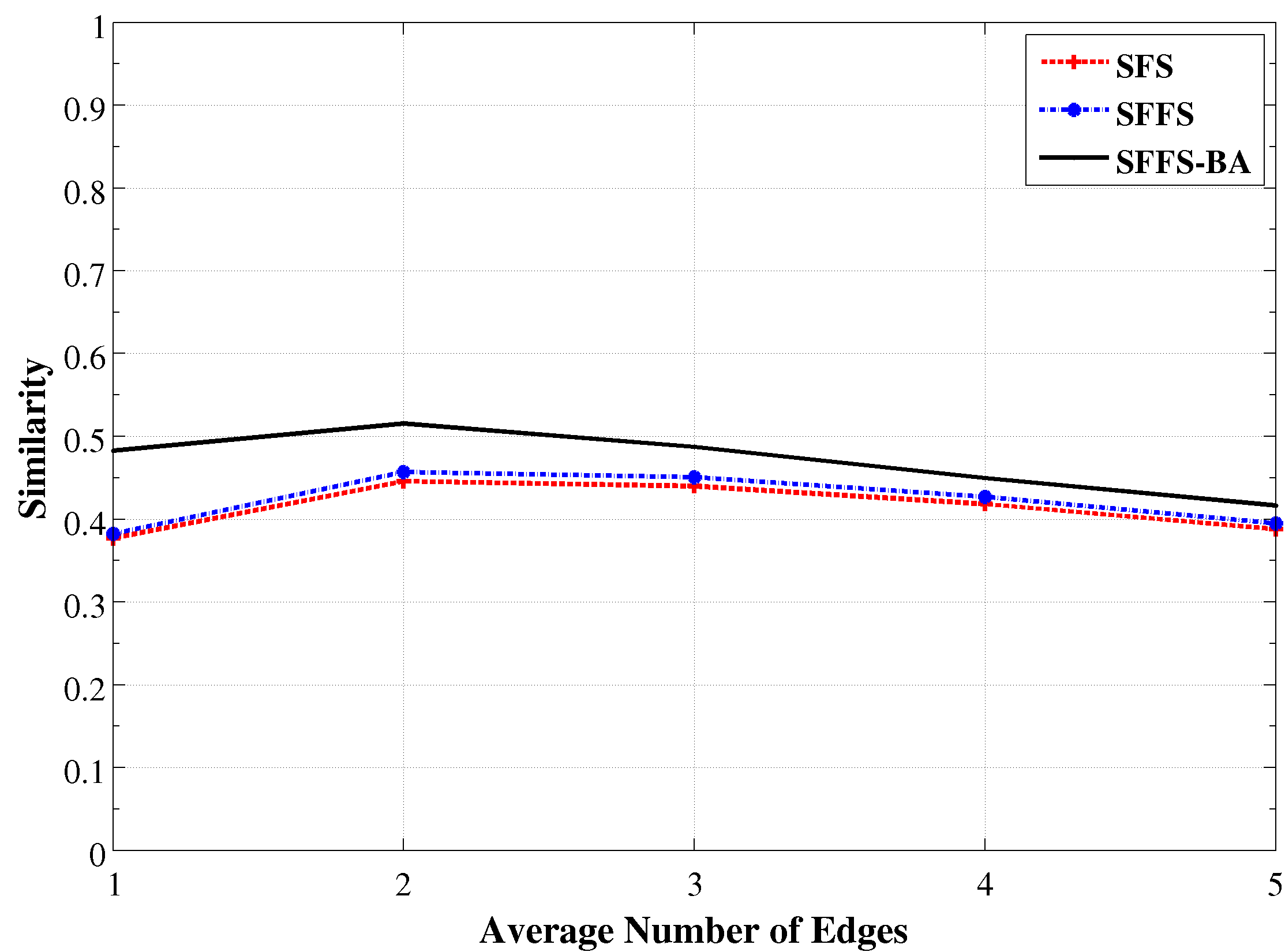} & \raisebox{0.31\linewidth}{(a) ER}\\
        \includegraphics[width=0.8\linewidth]{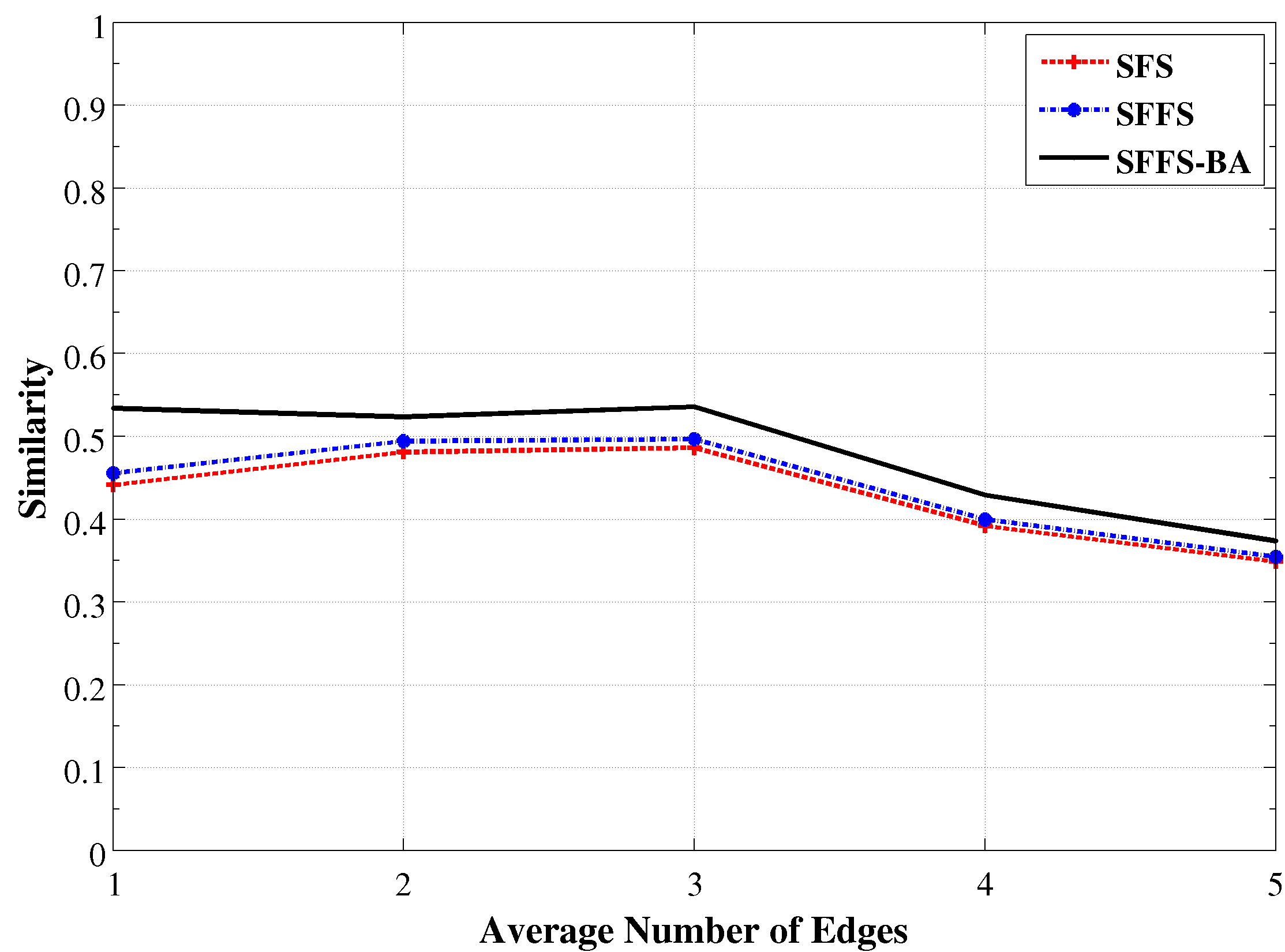} & \raisebox{0.31\linewidth}{(b) WS}\\
        \includegraphics[width=0.8\linewidth]{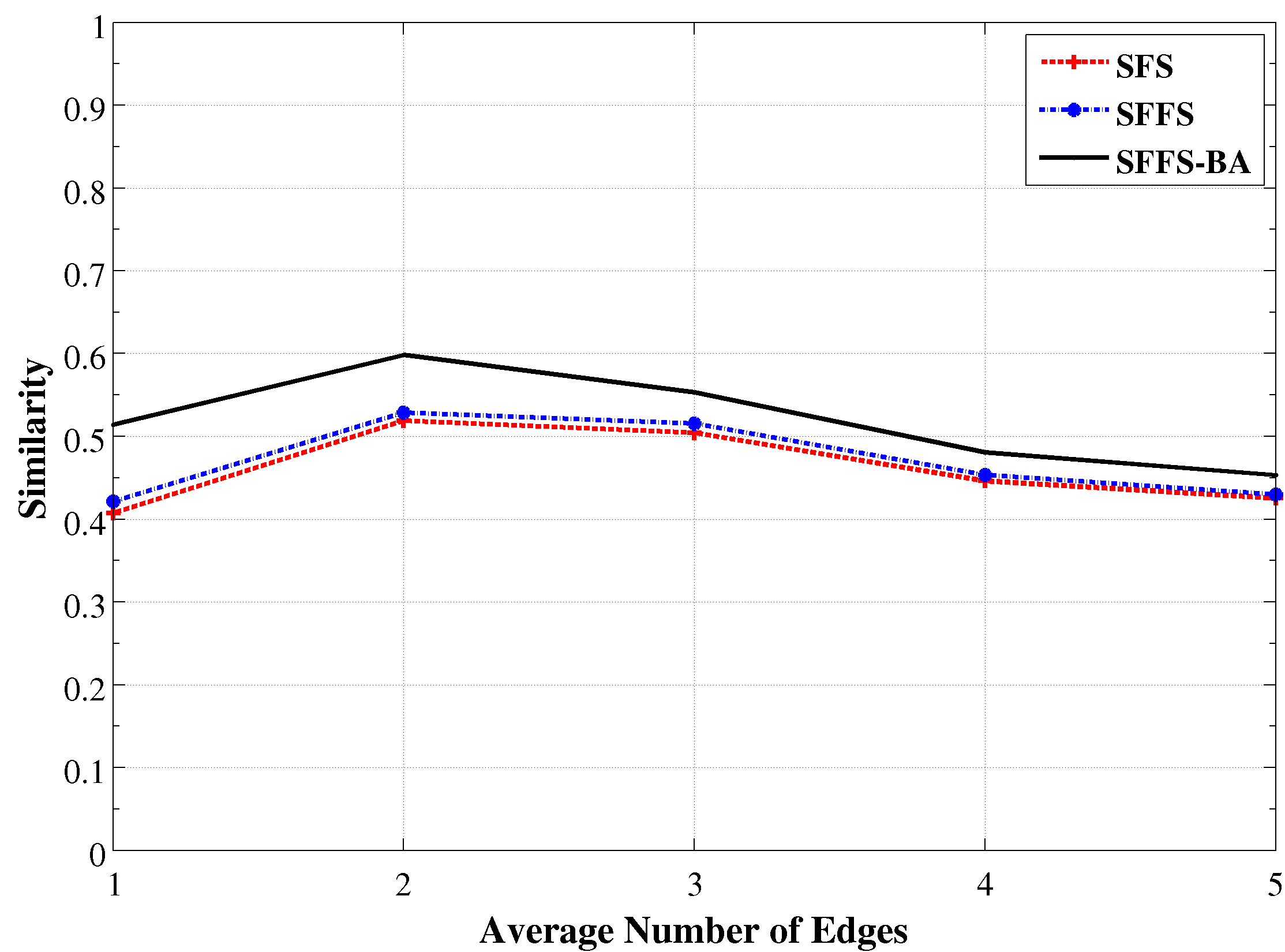} & \raisebox{0.31\linewidth}{(c) BA}\\
    \end{tabular}
    \caption{Similarity measure obtained by SFS, SFFS and SFFS-BA
            applied to infer network edges from different network complexities
            in terms of average degree $\langle k \rangle$.
            The similarity represents the mean over 50 executions for each network topology.}
    \label{fig:avgedges}
\end{figure}

The second experiment was performed in order to
compare the robustness of the methods by increasing
the complexity of the networks in terms of its
average degree $\langle k \rangle$.
Figure \ref{fig:avgedges} presents the average results for all
variations of signal size (number of time points).
It is possible to notice the similarity down-grade with the
increase of average degree $\langle k \rangle$ for the three
algorithms.
However, there was an improvement in results from $\langle k \rangle=1$
to $\langle k \rangle=2$ for ER and BA topologies.
This behavior can be explained by the fact that less complex networks
$\langle k \rangle=1$
have several genes with no predictor, but the inference methods tend
to find false positives, thus reducing its similarity ratio.

In this context, the SFFS-BA algorithm also outperforms the SFS and SFFS,
presenting a soft decrease of similarity with the increase of average
degree $\langle k \rangle$ for ER topology.
In the presence of a network structure as is the case of BA and WS topologies,
the decrease of similarity was less smooth, but even in these cases the SFFS-BA
presents better results.

With regard to the IMP genes with cardinality greater or equal than 3, it is
important to notice that the SFFS-BA tends to consider them if, at each step,
an individual predictor added to the subset has a prediction gain larger than
the predictors of the genes considered as sources (absence of predictors). The
tendency is that a moderately IMP set is detected if the number of samples
contained in the gene expression matrix is sufficient to estimate its joint
probability distributions. However, there are two situations in which it is
possible that SFFS-BA considers the target gene from an IMP set as source
gene. The first situation is the case in which the dimension of the IMP set is
excessive for the number of samples available, making the error estimation
of the joint probability distributions a crucial factor (the
number of parameter to be estimated grows exponentially as a function of the
cardinality). The second case refers to the strongly IMP sets where all its
properly contained subsets offer a very poor information gain with regard to
the target. This problem is inherent to the feature selection methods that
explore only a subspace of all possible solutions, as SFFS-BA does. The only
way to guarantee that IMP features are returned is through a exhaustive search
for the whole solution space.

\section{Conclusion}
\label{sec:conclusion}
This work presents an iterative floating search strategy for the inference of gene regulatory networks
by including the scale-free assumption as a prior information in the inference process.
Given the known limitations, our focus is the inclusion of prior knowledge on search methods.
In particular, by presenting a more suitable and efficient algorithm for the inference of GRNs from
temporal expression profiles, which presents a small number of samples and huge dimensionalities (genes).

The proposed algorithm is based on the assumption that several biological networks can be approximated 
by a scale-free topology. The presented method exploits this property by pruning the search space and
using a power law as a weight for reducing the search space.
In this context, the search space traversed by the SFFS-BA method
combines a breadth-first search when the number of combinations is small ($\langle k \rangle \leq 2$)
with a depth-first search when the number of combinations becomes explosive ($\langle k \rangle \geq 3$).

The experimental results show that the SFFS-BA provides better inference accuracy
than SFS and SFFS, when considering small signal sizes with 20-30
time points and also with large ones, with 100 time points.
In addition, the SFFS-BA was able to achieve $60\%$ of
similarity on network recovery after only $50$ observations from
a state space of size $2^{20}$, presenting very good results.

The SFFS-BA has also proved to be robust and stable, as SFS and SFFS,
when submitted to the increasing complexity of the networks in terms
of its average degree $\langle k \rangle$.
The robustness ans stability are important properties for the inference methods, 
even in the presence of some perturbations in the temporal signal, implied by the 
stochasticity in the application of transition functions.
Besides, the SFFS-BA showed better results than the SFS and SFFS.

A possible extension of the present work would be the inclusion
of the small-world (WS) \cite{ws1998} topology information in order to
guide the search process for the correct topology inference of these networks.
Also, we plan to apply this technique to infer GRNs from real data.
%
\section*{Acknowledgments}
This work was supported by FAPESP, CNPq and CAPES.
\bibliographystyle{IEEEtran}
\bibliography{bibliografia}
\end{document}